\documentclass[11pt,a4paper]{article}
\usepackage[hyperref]{AACL-IJCNLP2020}
\usepackage{times}
\usepackage{latexsym}

\usepackage{microtype}

\aclfinalcopy 

\usepackage{CJKutf8}
\usepackage{makecell}
\usepackage{multirow}
\usepackage{multicol}
\usepackage{graphicx}
\usepackage{caption}
\usepackage{subcaption}

\newcommand{\korean}[1]{\begin{CJK*}{UTF8}{mj}{#1}\end{CJK*}}

\title{An Empirical Study of Tokenization Strategies \\for Various Korean NLP Tasks}

\author{
Kyubyong Park$^{1,}$\thanks{*Equal contribution.} \space \space \space \space
Joohong Lee$^{2,*}$ \space \space \space \space
Seongbo Jang$^{2,3,*}$ \space \space \space \space
Dawoon Jung$^{2,*}$ \\

$^1$Kakao Brain, 
$^2$Pingpong AI Research, Scatter Lab, \\
$^3$Pohang University of Science and Technology \\

\texttt{kyubyong.park@kakaobrain.com} \\
\texttt{\{joohong,dawoon\}@scatterlab.co.kr} \\
\texttt{jang.sb@postech.ac.kr}
}

\date{}

\begin{document}
\maketitle
\begin{abstract}
Typically, tokenization is the very first step in most text processing works.
As a token serves as an atomic unit that embeds the contextual information of text, how to define a token plays a decisive role in the performance of a model.

Even though Byte Pair Encoding (BPE) has been considered the \emph{de facto} standard tokenization method due to its simplicity and universality, it still remains unclear whether BPE works best across all languages and tasks.
In this paper, we test several tokenization strategies in order to answer our primary research question, that is, ``What is the best tokenization strategy for Korean NLP tasks?''

Experimental results demonstrate that a hybrid approach of morphological segmentation followed by BPE works best in Korean to/from English machine translation and natural language understanding tasks such as KorNLI, KorSTS, NSMC, and PAWS-X.
As an exception, for KorQuAD, the Korean extension of SQuAD, 
BPE segmentation turns out to be the most effective.

Our code and pre-trained models are publicly available at \url{https://github.com/kakaobrain/kortok}.
\end{abstract}


\section{Introduction}
\label{sec:intro}
Tokenization is the very first step in most text processing works.
Not surprisingly, tremendous academic efforts have been made to find the best tokenization method for various NLP tasks.
For the past few years, Byte Pair Encoding (BPE) \citep{Gage:1994:NAD:177910.177914} has been considered the \emph{de facto
} standard tokenization technique since it was reintroduced by \citet{sennrich-etal-2016-improving}.
Besides the fact that BPE turns out to be very effective in the machine translation task, another important reason BPE has gained such popularity is that BPE is a data-driven statistical algorithm so it is independent of language.
However, it is still not clear whether BPE works best across all languages, irrespective of tasks.

In this paper we study various tokenization strategies for Korean, a language which is morphologically by far richer than English.
Concretely, we empirically examine what is the best tokenization strategy for Korean to English / English to Korean machine translation tasks, and natural language understanding (NLU) tasks---machine reading comprehension (MRC), natural language inference (NLI), semantic textual similarity (STS), sentiment analysis, and paraphrase identification. We are particularly interested in how complementary BPE and linguistically motivated segmentation are.

\begin{table*}
    \centering
    \begin{tabular}{p{5.5cm}p{7.5cm}}
        \Xhline{1.1pt}
        \bf{Tokenization} & \bf{Tokenized Sequence} \\
        \Xhline{1.1pt}
        Raw Text & \korean{나랑~쇼핑하자.}  \\ 
        \hline
        CV (\ref{jamo}) & \korean{ㄴ/ㅏ/ㄹ/ㅏ/ㅇ/$\star$/ㅅ/ㅛ/ㅍ/ㅣ/ㅇ/ㅎ/ㅏ/ㅈ/ㅏ/.}  \\
        Syllable (\ref{syl})& \korean{나/랑/$\star$/쇼/핑/하/자/.} \\
        Morpheme (\ref{mecab})& \korean{나/랑/$\star$/쇼핑/하/자/.}  \\
        Subword (\ref{bpe}) & \korean{\_나랑/\_쇼/핑하/자/.} \\
        Morpheme-aware Subword (\ref{mecab_bpe}) & \korean{\_나/\_랑/$\star$/\_쇼/핑/\_하/\_자/\_.} \\
        Word (\ref{word}) & \korean{나랑/쇼핑하자/.} \\
        \Xhline{1.1pt}
    \end{tabular}
    \caption{An input sentence \korean{나랑~쇼핑하자.} `Let's go shopping with me.' is differently tokenized depending on the various tokenization strategies. Slashes (/) are token separators.}
    \label{tab:ex_tok}
\end{table*}

\section{Background}
\label{sec:background}

\subsection{MeCab-ko: A Korean Morphological Analyzer}
\label{subsec:mecab-ko}
MeCab \citep{kudo2006mecab} is an open-source morphological analyzer based on Conditional Random Fields (CRFs).
It is originally designed for Japanese, but also serves generic purposes so it can be applied to other languages.
MeCab-ko\footnote{\url{https://bitbucket.org/eunjeon/mecab-ko}}, a Korean extension of MeCab, started from the idea that MeCab can be easily extended to the Korean language due to the close similarity between Japanese and Korean in terms of morphology or syntax.

MeCab-ko trained its model on the Sejong Corpus~\citep{kang200121st}, arguably the largest Korean corpus morphologically annotated by many experts, using MeCab. Ever since released in 2013, MeCab-ko has been widely used for many Korean NLP tasks due to its high accuracy and good usability. For example, the Workshop on Asian Translation (WAT) has adopted it as the official segmentation tool for evaluating Korean machine translation results since 2015. \citep{nakazawa-etal-2015-overview,nakazawa-etal-2016-overview,nakazawa-etal-2017-overview,nakazawa-etal-2018-overview,nakazawa-etal-2019-overview}.


\subsection{Byte Pair Encoding}
\label{subsec:bpe}

Byte  Pair  Encoding  (BPE)  is  a  simple  data  compression technique that iteratively replaces the most frequent pair of bytes in text with a single, unused byte~\citep{Gage:1994:NAD:177910.177914}. Since~\citet{sennrich-etal-2016-neural} successfully  applied  it  to  neural machine translation models, it has been regarded as the standard tokenization method across languages.

Korean is not an exception; \citet{park-etal-2019-knu} applied BPE to the Korean text in the Korean to Japanese task of WAT 2019 and ranked first. 
In addition, most recent Korean neural language models (e.g., KoBERT\footnote{\url{https://github.com/SKTBrain/KoBERT}}) used BPE to tokenize the training text.


\section{Related Work}
\label{sec:related-work}
There have been extensive studies about tokenization techniques for machine translation.
Several papers claimed that a hybrid of linguistically informed segmentation and a data-driven method like BPE or unigram language modeling performs the best for non-English languages. \citet{banerjee-bhattacharyya-2018-meaningless} combined an off-the-shelf morphological segmenter and BPE in Hindi and Bengali translations against English. \citet{tawfik-etal-2019-morphology} used a retrained version of linguistically motivated segmentation model along with statistical segmentation methods for Arabic. \citet{pinnis2017neural} adopted linguistic guidance to BPE for English-Latvian translation.
Particularly~\citep{park2019korean-subword} is close to ours, but their main focus is on preprocessing techniques for neural machine translation like parallel corpus filtering rather than on tokenization strategies per se.

Compared with the tokenization studies for machine translation, those for NLU tasks have gained less attention.
Among them is \citet{bostrom2020byte}, which compared the fine-tuning task performance of BERT \cite{devlin-etal-2019-bert} pre-trained with BPE and unigram language modeling. \citet{moon-okazaki-2020-jamo} proposed a novel encoding method for Korean and showed its efficiency in vocabulary compression with a few Korean NLU datasets.


\section{Tokenization Strategies}
\label{sec:tokenization_strategies}

We introduce assorted Korean tokenization strategies arranged from the smallest to the largest unit. Each of them induces different tokenization results, as illustrated in Table \ref{tab:ex_tok}.

\subsection{Consonant and Vowel (CV)}
\label{jamo}

In Hangul, the standard Korean writing system, consonants and vowels, called \emph{Jamo} in Korean, corresponding to Latin letters are assembled to form a syllable character. 
For example, a Hangul consonant \korean{ㅎ} /h/ (U+314E) is combined with a vowel \korean{ㅏ} /a/ (U+314F) to make a syllable character \korean{하} /ha/ (U+558).
Readers who are not familiar with such a mechanism can think of Jamo and syllables as atoms and molecules respectively.
As a molecule \texttt{H\textsubscript{2}O} can be decomposed into two \texttt{H} atoms and an \texttt{O} atom, a syllable \korean{하} /ha/ can be decomposed into its constituent consonant \korean{ㅎ} /h/ and vowel \korean{ㅏ} /a/.
The first syllable \korean{나} /na/ of the raw text in Table \ref{tab:ex_tok} is tokenized into \korean{ㄴ} /n/ and \korean{ㅏ} /a/, and the second syllable \korean{랑} /lang/ is tokenized into \korean{ㄹ} /l/, \korean{ㅏ} /a/, and \korean{ㅇ} /ng/, and so on.
A whitespace is replaced by a special symbol $\star$.

\subsection{Syllable}
\label{syl}
We can tokenize a sentence at the syllable level.
A whitespace is replaced by the special symbol $\star$.

\subsection{Morpheme}
\label{mecab}
MeCab-ko provides a convenient tokenization option in the command line interface\footnote{\% mecab -O wakati}.
For example, it returns \texttt{A}, \texttt{B}, and \texttt{C} given an input text \texttt{AB C}, where \texttt{A}-\texttt{C} represent morphemes.
Note that the original space between \texttt{AB} and \texttt{C} is missing in the output token list.
Accordingly, it is NOT possible to recover the original text from the tokenized result.

This can be problematic in some tasks that require us to restore the input text such as machine translation whose target language is Korean, or machine reading comprehension where we are expected to suggest a certain phrase in the given text as the answer.

For this reason, we insert a special token $\star$ (U+2B51) to the original whitespace position.
As a result, in the above example, the tokenized sequence looks like \texttt{A}, \texttt{B}, $\star$, and \texttt{D}.

\subsection{Subword}
\label{bpe}

We learn and apply BPE using the SentencePiece \citep{kudo-richardson-2018-sentencepiece} library.
It prepends `\_' (U+2581) to every word to mark the original whitespace, then tokenizes text into subword pieces.
As seen in Table \ref{tab:ex_tok}, \korean{나랑~쇼핑하자.} can be split into \korean{\_나랑}, \korean{\_쇼}, \korean{핑하}, \korean{자}, and . (period).

\subsection{Morpheme-aware Subword}
\label{mecab_bpe}
Motivated by the combined methods of data- and linguistically-driven approaches~\citep{banerjee-bhattacharyya-2018-meaningless,park2019korean-subword,pinnis2017neural,tawfik-etal-2019-morphology}, we apply MeCab-ko and BPE in sequence to make morpheme-aware subwords.
According to this strategy, since BPE is applied \emph{after} the original text is split into morphemes, tokens spanning multiple morphemes (e.g., \korean{핑하} in the Section \ref{bpe}) are not generated. Instead, the BPE algorithm further segments morphemes into frequent pieces.

\subsection{Word}
\label{word}

We can simply split text by whitespaces. Note that punctuation marks are split into separate tokens.
Check that \korean{나랑~쇼핑하자.} is tokenized into \korean{나랑}, \korean{쇼핑하자} and . (period) in Table \ref{tab:ex_tok}.


\section{Experiments}
\label{exp}

\setlength{\tabcolsep}{3.5pt}
\begin{table}
    \centering
    \begin{tabular}{ccccc}
    \Xhline{1.1pt}
    \makecell{\bf{Lang}\\\bf{Pair}} & \makecell{\bf{Vocab}\\\bf{Size}} & \makecell{\bf{Korean BPE}\\\bf{Training Data}} & \bf{Dev} & \bf{Test} \\
    \Xhline{1.1pt}
    \multirow{2}{*}{Ko-En}
    & \multirow{2}{*}{32K}&AI Hub (130MB) & 35.79 & 36.06 \\
    &  & Wiki (613MB)    & \bf{39.05} & \bf{38.69} \\
    
    \hline
    \multirow{2}{*}{En-Ko} 
    
    & \multirow{2}{*}{32K}&AI Hub (130MB) & \bf{37.19} &\bf{36.98} \\
    & &Wiki (613MB)    & 37.11 & \bf{36.98} \\
    \Xhline{1.1pt}
    \end{tabular}
    \caption{BLEU scores of Korean to English (Ko-En) and English to Korean (En-Ko) translation models with different BPE training data. Note that the English sentences are tokenized using a 32K BPE model trained on the English Wiki.}
    \label{tab:bpe-data}
\end{table}

\begin{table*}
\setlength{\tabcolsep}{6pt}
\centering
\begin{tabular}{lccccccc}
\Xhline{1.1pt}
\multirow{2}{*}{\bf{Tokenization}}
& \multirow{2}{*}{\makecell{\bf{Vocab Size}}}
& \multicolumn{2}{c}{\bf{Ko-En}}
& \multicolumn{2}{c}{\bf{En-Ko}}
& \multirow{2}{*}{\makecell{\bf{OOV Rate (\%)}}}
& \multirow{2}{*}{\makecell{\bf{Avg. Length}}} \\
\cline{3-6}
& & \bf{Dev} & \bf{Test} & \bf{Dev} & \bf{Test} &  &  \\

\Xhline{1.1pt}
CV
& 166         & 39.11         & 38.56   &36.52&36.45    & 0.02  & 142.75 \\
\hline
Syllable
& 2K          & 39.30        & 38.75  &38.64&38.45    & 0.06  & 69.20  \\
\hline
\multirow{4}{*}{Morpheme}
& 8K          & 31.59        & 31.24 &32.44&32.19     & 7.51 & 49.19  \\
& 16K         & 34.38        & 33.80  &35.74&35.52    & 4.67  & 49.19  \\
& 32K         & 36.19        & 35.74   &36.51&36.12   & 2.72  & 49.19  \\
& 64K         & \underline{37.88}        & \underline{37.37}  & \underline{37.51}&\underline{37.03}   & 1.40  & 49.19  \\
\hline

\multirow{5}{*}{Subword}
& 4K          & 39.18        & 38.75   &\underline{38.31}         & \underline{38.18}   & 0.07  & 48.02  \\
& 8K          & 39.16        & 38.75 &38.09&37.94     & 0.08  & 38.44  \\
& 16K         & \underline{39.22}        & \underline{38.77} &37.64&37.34     & 0.10  & 33.69  \\
& 32K         & 39.05        & 38.69   &37.11&36.98   & 0.11  & 30.21  \\
& 64K         & 37.02        & 36.46  &35.77&35.64    & 0.12  & 27.50  \\
\hline
\multirow{5}{*}{\makecell[l]{Morpheme-aware\\Subword}}
& 4K          & 39.41        & 38.95  &39.29&39.13    & 0.06  & 65.17  \\
& 8K          & 39.42        & 39.06   &39.78&39.61   & 0.06  & 56.79  \\
& 16K         & 39.84        & 39.41  &40.23& 40.04   & 0.07  & 53.30  \\
& 32K         & \bf{41.00}   & \bf{40.34} &\bf{40.43}&\bf{40.41}& 0.07  & 51.38  \\
& 64K         & 39.62        & 39.34   &38.63&38.42   & 0.07  & 50.27  \\
\hline
\multirow{1}{*}{Word}
& 64K         & 7.04         & 7.07  &18.68&18.42     & 26.20  & 18.96  \\
\Xhline{1.1pt}
\end{tabular}
\caption{BLEU scores of Korean to English (\textbf{Ko-En}) and English to Korean (\textbf{En-Ko}) translation models of various tokenization strategies. Note that we use an 32K Subword model for English for all of them. The OOV rate values in the table are obtained from the test set, but there is no meaningful difference between the test and the dev set in terms of the OOV rate. The best BLEU scores in each column (global) and group (local) are bold-faced and underlined, respectively.}
\label{tab:mt}
\end{table*}

\subsection{Korean to/from English Machine Translation}
\label{nmt}

\subsubsection{Dataset}
\label{data}
To date, there have yet been few open source benchmark datasets for Korean-English machine translation, not to mention that Korean is not in the language list of WMT\footnote{\url{https://www.aclweb.org/anthology/venues/wmt}} or IWSLT\footnote{\url{http://iwslt.org/doku.php?id=start}}.
\citet{park2019korean-subword} used OpenSubtitles \citep{lison-tiedemann-2016-opensubtitles2016}, a collection of crowd-sourced movie subtitles across 65 different languages, for English to Korean translation, but they are too noisy to serve as a translation benchmark dataset.\footnote{\citet{park2019korean-subword} reported BLEU scores of 7-12.}

Recently, a Korean-English parallel corpus was publicly released by AI Hub\footnote{\url{http://www.aihub.or.kr/aidata/87}}, which was gathered from various sources such as news, government web sites, legal documents, etc.
We download the news data, which amount to 800K sentence pairs, and randomly split them into 784K (train), 8K (dev), and 8K (test).

\subsubsection{BPE Modeling}
\label{subsec:bpe_modeling}

Prior to training, we do simple preliminary experiments to decide which dataset to use for learning BPE.

There are two choices: AI Hub news training data and open source large text such as Wiki.
AI Hub training data is relatively small in size (130 MB), but can be optimal as its lexical distribution will be close to that of the test data, considering both of them are from the same source.
On the other hand, Wiki is larger, but it is not news per se, so can be not as appropriate as AI Hub data for BPE modeling.

To investigate this, first we train a 32K Korean BPE model (\textbf{A}) using SentencePiece with the Korean sentences in the AI Hub training data. 
Then we download the latest Wikipedia Korean\footnote{\url{https://dumps.wikimedia.org/kowiki}}/English\footnote{\url{https://dumps.wikimedia.org/enwiki}} dumps, and extract plain texts using WikiExtractor \footnote{\url{https://github.com/attardi/wikiextractor}}.
Next, we make 32K BPE models for Korean (\textbf{B}) and English (\textbf{C}) with them.
Finally, we train Korean to English (Ko-En) and English to Korean (En-Ko) translation models on the AI Hub training data with the two different Korean BPE models (\textbf{A, B}). The training details are explained in Section \ref{subsec:training}. For comparison, we use the same English BPE model (\textbf{C}) for both.

The results are shown in Table \ref{tab:bpe-data}.
For Ko-En translation, the Wiki-based BPE model performs better in both dev and test sets by 2-3 points.
For En-Ko translation, there is no practical difference in performance between the Wiki and AI Hub-based models.
It is also worth considering the BPE models are used for NLU tasks as well as machine translation.
All things taken together, we opt for the Wiki-based BPE model.

\subsubsection{Training}
\label{subsec:training}
We test the tokenization strategies in Section \ref{sec:tokenization_strategies} with various vocabulary sizes on the AI Hub news dataset.

We use the Transformer \cite{vaswani2017attention}, the state-of-the-art model for neural machine translation.
We mostly follow the base model configuration: 6 blocks of 512-2048 units with 8 attention heads.
We run all of our experiments using \textsc{FAIRSEQ} \footnote{\url{https://github.com/pytorch/fairseq}} \cite{ott-etal-2019-fairseq}, a PyTorch based deep learning library for sequence to sequence models.

Each model is trained using a Tesla V100 GPU with batch size 128, dropout rate 0.3, label smoothing 0.1, and the Adam \citep{kingma2015adam} optimizer. We set the learning rate to 5e-4 with the inverse square-root schedule.
We train all models for 50 epochs and save the checkpoint files at every epoch.

\subsubsection{Results}
\label{subsec:results}

\begin{table}
\small
\setlength{\tabcolsep}{7pt}
    \centering
    \begin{tabular}{crr}
    \Xhline{1.1pt}
    \bf{Vocab Size} & \bf{\# Tokens} & \makecell{\bf{\# Tokens Spanning}\\ \bf{Morpheme Boundaries}}\\
    \hline

    4K  & 387,088 & 25,458 (6.58\%)  \\
    8K  & 309,360 & 50,029 (16.17\%) \\
    16K & 271,334 & 62,861 (23.17\%) \\
    32K & 242,736 & 73,609 (30.26\%) \\
    64K & 221,530 & 82,324 (37.16\%) \\

    \Xhline{1.1pt}
    \end{tabular}
    \caption{Number of tokens spanning morpheme boundaries in Subword models.}
    \label{tab:token_boundaries}
\end{table}

After all training stages are finished, we evaluate the saved checkpoint files of each model on the dev set to find the best one, which is subsequently used for the final test.
In Table \ref{tab:mt} we report BLEU scores on both the dev and test sets using the Moses\footnote{\url{http://www.statmt.org/moses}} \texttt{multi-bleu.perl} script.
Following WAT 2019 \citep{nakazawa-etal-2019-overview}, Moses tokenizer and MeCab-ko are used for tokenizing the evaluation data.

For both Ko-En and En-Ko, overall, the Subword models (35.64-39.22) and the Syllable models (38.45-39.30) are superior to the Morpheme models (31.59-37.37) or the Word models (7.04-18.42) in performance. 
It is highly likely to come from the lower OOV rates of the Subword models (0.07-0.12) and the Syllable models (0.06) compared to those of the Morpheme models (1.40-7.51) and the Word models (26.20).
While BPE tends to split rare words into subword pieces, MeCab-ko is ignorant of statistics so it splits words into morphemes by linguistic knowledge instead. That the Morpheme and Word models generate many OOVs suggests Korean has so large types of morphemes or word forms that even 64K vocabulary is not enough to cover them all.

CV models are tiny in vocabulary size (166) so they show the lowest OOV rate (0.02).
However, their performance is not as good as the Syllable or Subword models.
We speculate this is because a single consonant or vowel must bear too much contextual information in the CV models.

Morpheme-aware Subword 32K models achieve the best BLEU scores.
Each Subword model, as shown in Table \ref{tab:token_boundaries}, contains 6-37\% of tokens spanning morpheme boundaries in the test set, which implies that subword segmentation by BPE is not optimal and morpheme boundaries are meaningful in tokenization.

To sum up, morpheme-aware subword tokenization that makes the best use of linguistic knowledge and statistical information is the best for Korean machine translation.

\subsection{Korean Natural Language Understanding Tasks}
\label{subsec:nlu}
Large pre-trained language models have proven their effectiveness in many downstream tasks \citep{peters-etal-2018-deep,devlin-etal-2019-bert,liu2019roberta}.
We pre-train  BERT~\citep{devlin-etal-2019-bert} models with various tokenization strategies, and fine-tune them on five different Korean NLU tasks.

\subsubsection{Machine Reading Comprehension: KorQuAD 1.0 Dataset}
\label{subsec:korquad}

The KorQuAD 1.0 dataset \citep{lim2019korquad1} is a Korean adaptation of SQuAD 1.0 \citep{rajpurkar-etal-2016-squad}, a popular reading comprehension dataset.
KorQuAD 1.0 consists of 10,645 passages and their paired 66,181 questions (60,407 for training + 5,774 for development\footnote{The test dataset is not included.}). Like SQuAD 1.0, KorQuAD 1.0 involves answering a question given a passage. The answer must be a phrase within the passage.

\setlength{\tabcolsep}{2pt}
\begin{table}
\small
    \centering
    \begin{tabular}{cccccc}
    \Xhline{1.1pt}
    \makecell{\bf{Hyper-}\\\bf{param}} & \bf{KorQuAD} & \bf{KorNLI} & \bf{KorSTS} & \bf{NSMC} & \bf{PAWS} \\
    \hline
    Epoch               & 5 & 3    & 5    & 3    & 5 \\
    Batch          & 16 & 64   & 64   & 64   & 64 \\
    $\eta$       & 5e-5 & 1e-4 & 5e-5 & 5e-5 & 1e-4 \\
    Dropout     & 0.1 & 0.1 & 0.1 & 0.1 & 0.1 \\
    Warm-up      & 0.1 & 0.1 & 0.1 & 0.1 & 0.1 \\
    Max Seq.$^\dagger$     & 128 & 128 & 128 & 128 & 128 \\
    \Xhline{1.1pt}
    \end{tabular}
    \caption{Fine-tuning hyper-parameters for NLU tasks. $\eta$: learning rate. $\dagger$: Max sequence length is 256 for CV models in all tasks.}
    \label{tab:hyperparameter}
\end{table}

\begin{table*}
\centering
\setlength{\tabcolsep}{4pt}
\begin{tabular}{lcccccccccccc}
\Xhline{1.1pt}
\multirow{2}{*}{\bf{Tokenization}} 
& \multirow{2}{*}{\makecell{\bf{Vocab}\\\bf{Size}}}
& \bf{KorQuAD}
& \multicolumn{2}{c}{\bf{KorNLI}} 
& \multicolumn{2}{c}{\bf{KorSTS}}  
& \multicolumn{2}{c}{\bf{NSMC}}  
& \multicolumn{2}{c}{\bf{PAWS-X}} \\
\cline{3-11}
& & \bf{Dev (EM/F1)} & \bf{Dev} & \bf{Test} & \bf{Dev} & \bf{Test} & \bf{Dev} & \bf{Test} & \bf{Dev} & \bf{Test} \\
\Xhline{1.1pt}
CV
& 166         & 59.66 / 73.91 & 70.60 & 71.20 & 77.22 & 71.47 & 87.97 & 87.89 & 58.00 & 55.20 \\
\hline
Syllable
& 2K          & 69.10 / 83.29 & 73.98 & 73.47 & 82.70 & 75.86 & 88.94 & 89.07 & 68.65 & 67.20 \\
\hline
\multirow{2}{*}{Morpheme}
& 32K         & 68.05 / 83.82 & 74.86 & 74.37 & 82.37 & 76.83 & 87.87 & 88.04 & 69.30 & 67.20 \\
& 64K         & \underline{70.68} / \underline{85.25} & \underline{75.06} & \underline{75.69} & \underline{83.21} & \underline{77.38} & \underline{88.72} & \underline{88.88} & \underline{73.40} & \underline{68.65} \\
\hline
\multirow{5}{*}{Subword}
& 4K          & 71.48 / 83.11 & 74.38 & 74.03 & 83.37 & 76.80 & 89.08 & 89.30 & 72.00 & 69.60 \\
& 8K          & 72.91 / 85.11 & 74.18 & 74.65 & 83.23 & 76.42 & 89.08 & 89.19 & 73.45 & 69.00 \\
& 16K         & 73.42 / 85.75 & 74.46 & \underline{75.15} & 83.30 & 76.41 & 88.89 & 88.88 & 73.40 & 70.70 \\
& 32K         & \textbf{74.04} / 86.30 & \underline{74.74} & 74.29 & 83.02 & 77.01 & \underline{89.39} & \underline{89.38} & 74.05 & 70.95 \\
& 64K         & \textbf{74.04} / \textbf{86.66} & 73.73 & 74.55 & \underline{83.52} & \underline{77.47} & 88.80 & 89.19 & \underline{75.85} & \underline{72.10} \\
\hline

\multirow{5}{*}{\makecell[l]{Morpheme-aware\\Subword}}
& 4K          & 67.53 / 81.93 & 73.53 & 73.45 & 83.34 & 76.03 & 88.93 & 89.32 & 69.75 & 67.45 \\
& 8K          & 70.90 / 84.57 & 74.14 & 73.95 & 83.71 & 76.07 & 89.37 & 89.29 & 73.40 & 71.30 \\
& 16K         & 69.47 / 83.36 & 75.02 & 74.99 & 83.22 & 76.59 & 89.33 & 89.41 & 75.05 & 71.70 \\
& 32K         & \underline{72.65} / \underline{86.35} & 74.10 & 75.13 & 83.65 & \bf{78.11} & 89.53 & 89.65 & 74.60 & 71.60 \\
& 64K         & 69.48 / 83.73 & \bf{76.39} & \bf{76.61} & \bf{84.29} & 76.78 & \bf{89.82} & \bf{89.66} & \bf{76.15} & \bf{74.00} \\
\hline
Word
& 64K         & 1.54 / 8.86 & 64.06 & 65.83 & 69.00 & 60.41 & 70.10 & 70.58 & 58.25 & 55.30 \\
\Xhline{1.1pt}
\end{tabular}
\caption{Performance of various models on several Korean natural language understanding tasks. The evaluation metrics are as follows: KorQuAD: Exact Match/F1, KorNLI: accuracy (\%), KorSTS: 100 $\times$ Spearman correlation, NSMC: accuracy (\%), PAWS-X: accuracy (\%). The best scores in each column (global) and group (local) are bold-faced and underlined, respectively.}
\label{tab:nlu}
\end{table*}

\subsubsection{Natural Language Inference: KorNLI Dataset}
\label{subsec:kornli}
The KorNLI Dataset \citep{ham2020kornli} is a Korean NLI dataset sourced from three different NLI datasets: SNLI \citep{bowman-etal-2015-large}, MNLI \citep{williams2018broad}, and XNLI \citep{conneau2018xnli}. 

It is composed of 950,354 sentence pairs: 942,854 for training, 2,490 for development, and 5,010 for test. A model receives a pair of sentences---a premise and a hypothesis---and classifies their relationship into one out of three categories: \emph{entailment}, \emph{contradiction}, and \emph{neutral}.

\subsubsection{Semantic Textual Similarity: KorSTS Dataset}
\label{subsec:korsts}
The KorSTS Dataset \citep{ham2020kornli} is a Korean STS dataset translated from the STS-B dataset \citep{cer2017semeval}. 
It comprises 8,628 sentence pairs---5,749 for training, 1,500 for development, and 1,379 for test. The task assesses the gradations of semantic similarity between two sentences with a scale from 0 to 5.

\subsubsection{Sentiment Analysis: NSMC Dataset}
\label{subsec:nsmc}
NSMC\footnote{\url{https://github.com/e9t/nsmc}} is a movie review dataset scraped from Naver Movies\texttrademark. 
It consists of 200K samples of which 150K are the training set and the rest 50K are the test set.
Each sample is labeled with 0 (negative) or 1 (positive).
We hold out 10 percent of the training data for development.

\subsubsection{Paraphrase Identification: PAWS-X Dataset}
\label{subsec:paws}

The PAWS-X dataset \citep{yang-etal-2019-paws} is a challenging paraphrase identification dataset in six languages including Korean. 
The Korean portion amounts to 53,338 sentence pairs (49,410 for training, 1,965 for development, and 1,972 for test).
Like the NSMC dataset, each sentence pair is annotated with either 0 (negative) or 1 (positive).

\begin{figure*}
    \centering
    \begin{subfigure}[b]{.495\textwidth}
        \centering
        \includegraphics[width=\textwidth]{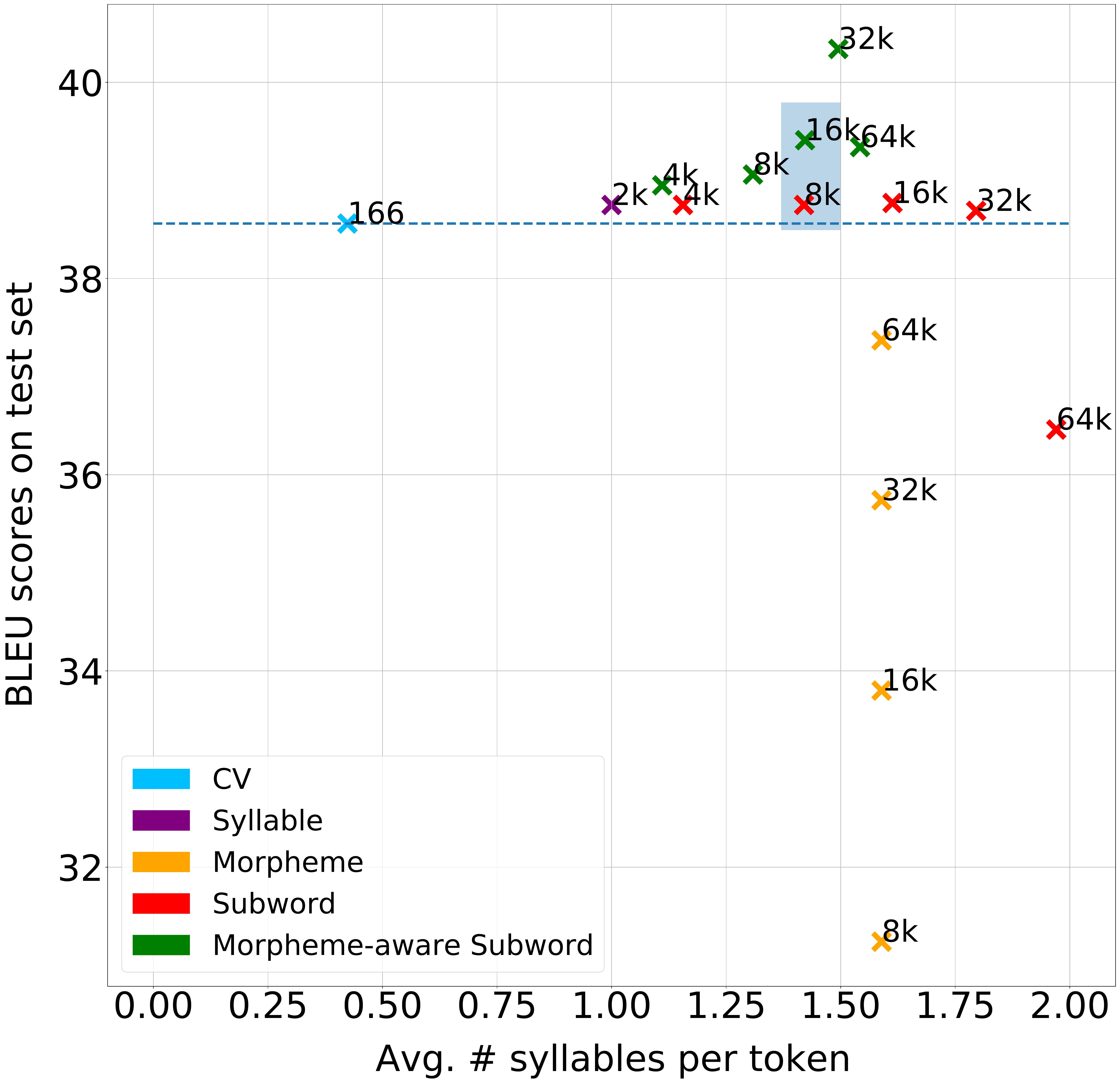}
        \caption{Ko-En}
        \label{fig:avg_syllabel_length_ko-en}
    \end{subfigure}
    \hfill
    \begin{subfigure}[b]{.495\textwidth}
        \centering
        \includegraphics[width=\textwidth]{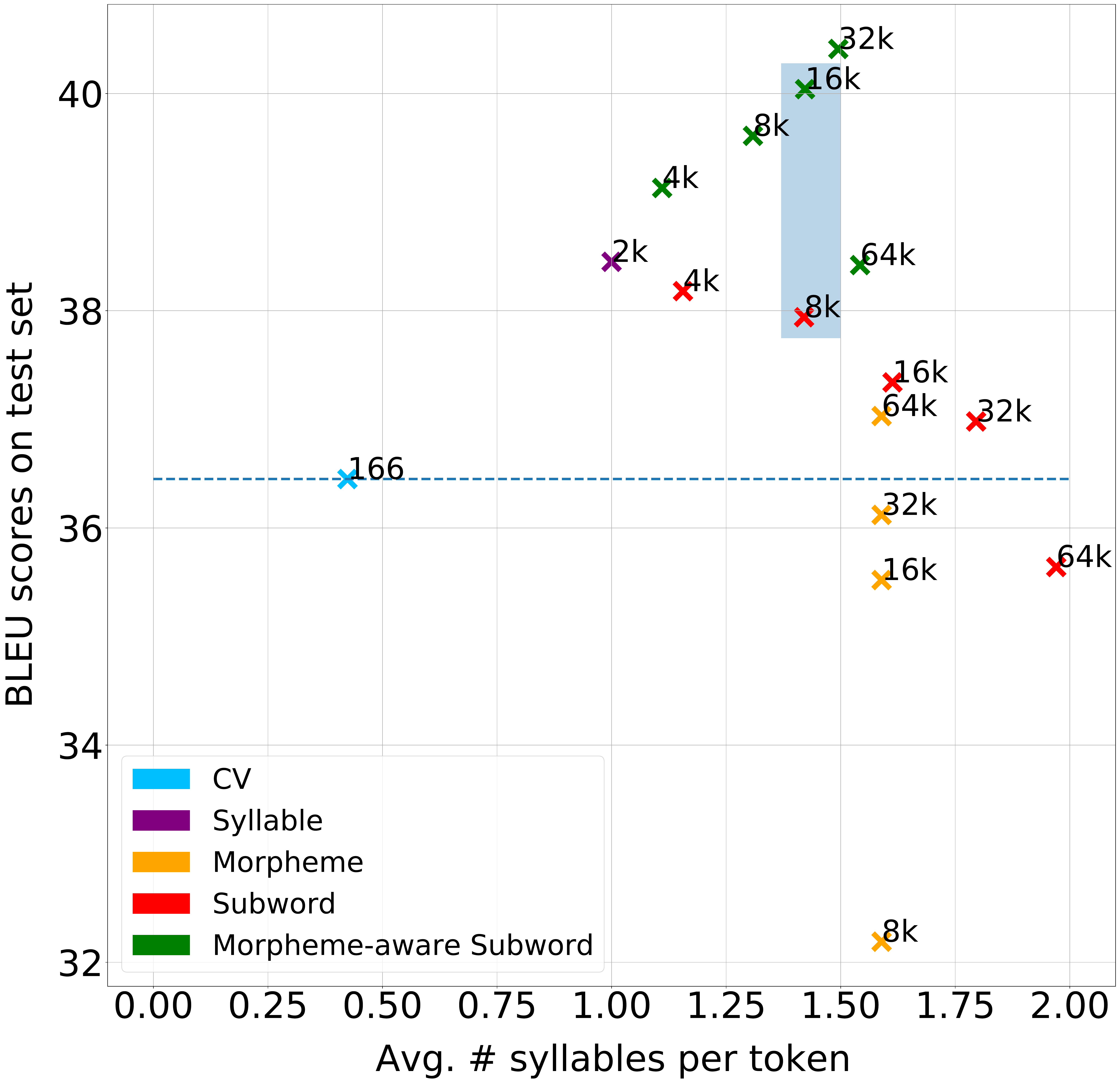}
        \caption{En-Ko}
        \label{fig:avg_syllable_length_en-ko}
    \end{subfigure}
    \caption{Translation performance over the average number of syllables per token}
    \label{fig:avg_syllable_length}
\end{figure*}

For each tokenization strategy, we pre-train a BERT-Base model on a large corpus and fine-tune it on the training sets of the five NLU tasks independently.

\noindent
\textbf{Pre-training.} 
Because the Korean Wiki corpus is not enough in volume, 640 MB, for the pre-training purpose, we additionally download the recent dump of Namuwiki\footnote{\url{http://dump.thewiki.kr}}, a Korean Wiki, and extract plain texts using Namu Wiki Extractor\footnote{\url{https://github.com/jonghwanhyeon/namu-wiki-extractor}}.
On the resulting Namuwiki corpus (5.5 GB) along with the Wiki corpus (640 MB), pre-training is performed with a Cloud TPU v3-8 for 1M steps using the official BERT training code\footnote{\url{https://github.com/google-research/bert}}, which is based on TensorFlow. We set the training hyper-parameters of all models as follows: batch size = 1024, max sequence length = 128, optimizer = AdamW \cite{loshchilov2019decoupled}, learning rate = 5e-5, warm-up steps = 10K.

\noindent
\textbf{Fine-tuning.}
After converting each of the pre-trained models in TensorFlow into PyTorch, we fine-tune it using HuggingFace Transformers\footnote{\url{https://github.com/huggingface/transformers}} \citep{Wolf2019HuggingFacesTS}.
The hyper-parameters for each task are shown in Table \ref{tab:hyperparameter}.

\subsubsection{Results}
\label{subsec:nlu_results}

In Table \ref{tab:nlu} we report the evaluation results of the various models on the dev and test sets.
Since KorQuAD lacks the test set, we report the results on the dev set only.

As for KorQuAD, Subword 64K models achieve the highest Exact Match (EM) and F1 scores. The scores in the Subword and Morpheme models increase monotonically as the vocabulary size grows. On the other hand, the 32K models outperform the others in the Morpheme-aware Subword models; no clear correlation is found between performance and vocabulary sizes in them.

For all the other four tasks, Morpheme-aware Subword 64K models show the best scores. One noteworthy phenomenon is that the scores tend to increase as the vocabulary size grows across the tokenization groups.
This is discordant with the machine translation results in Section \ref{subsec:results}, where a larger vocabulary size does not guarantee better performance for the Subword and Morpheme-aware Subword models.


\section{Discussion}
\label{sec:discussion}

We further examine which factors with respect to tokenization affect the Ko-En and En-Ko translation performance.

\subsection{Token Length}

Because tokenization involves splitting a text into shorter segments, we find it important to figure out how much information each segment bears.
To this end, based on the assumption that the longer a text is, the more information it is likely to have, we plot the BLEU scores by the average number of syllables per Korean token in the translation test sets in Figure \ref{fig:avg_syllable_length}.

The BLEU scores of the subword models---Syllable, Morpheme, Subword, and Morpheme-aware Subword---are mostly higher than those of the CV models, which are plotted as dotted lines. In particular, the Syllable, Subword, and Morpheme-aware Subword models between 1.00 and 1.50 show the best scores both in Ko-En and in En-Ko.
When a token has more than 1.5 syllables on average, the scores begin to decrease, and the Word models which has more than 2.5 syllables in a token performs the worst (7.07 for Ko-En and 18.42 for En-Ko). Note that they are not in the figures due to space constraints.

\subsection{Linguistic Awareness}
Obviously token length is not the only key factor in tokenization strategies.
Let us compare the Morpheme-aware Subword 16K  models (green markers) and Subword 8K models (red markers) in the shaded regions in Figure  \ref{fig:avg_syllable_length}. 
Although they have the same average token length around 1.4, the Morpheme-aware Subword models outperform the Subword models. 
We believe this is evidence to support that linguistic awareness is another important factor in Korean tokenization strategies for machine translation.

\subsection{Under-trained Tokens}
In section \ref{subsec:results}, we pointed out high OOV rates are highly likely to degrade the performance of Morpheme models. It is also worth noting that in Figure \ref{fig:avg_syllable_length} as most of the orange markers denoting Morpheme models are below the dotted lines. 

\begin{figure}
    \centering
    \includegraphics[width=\linewidth]{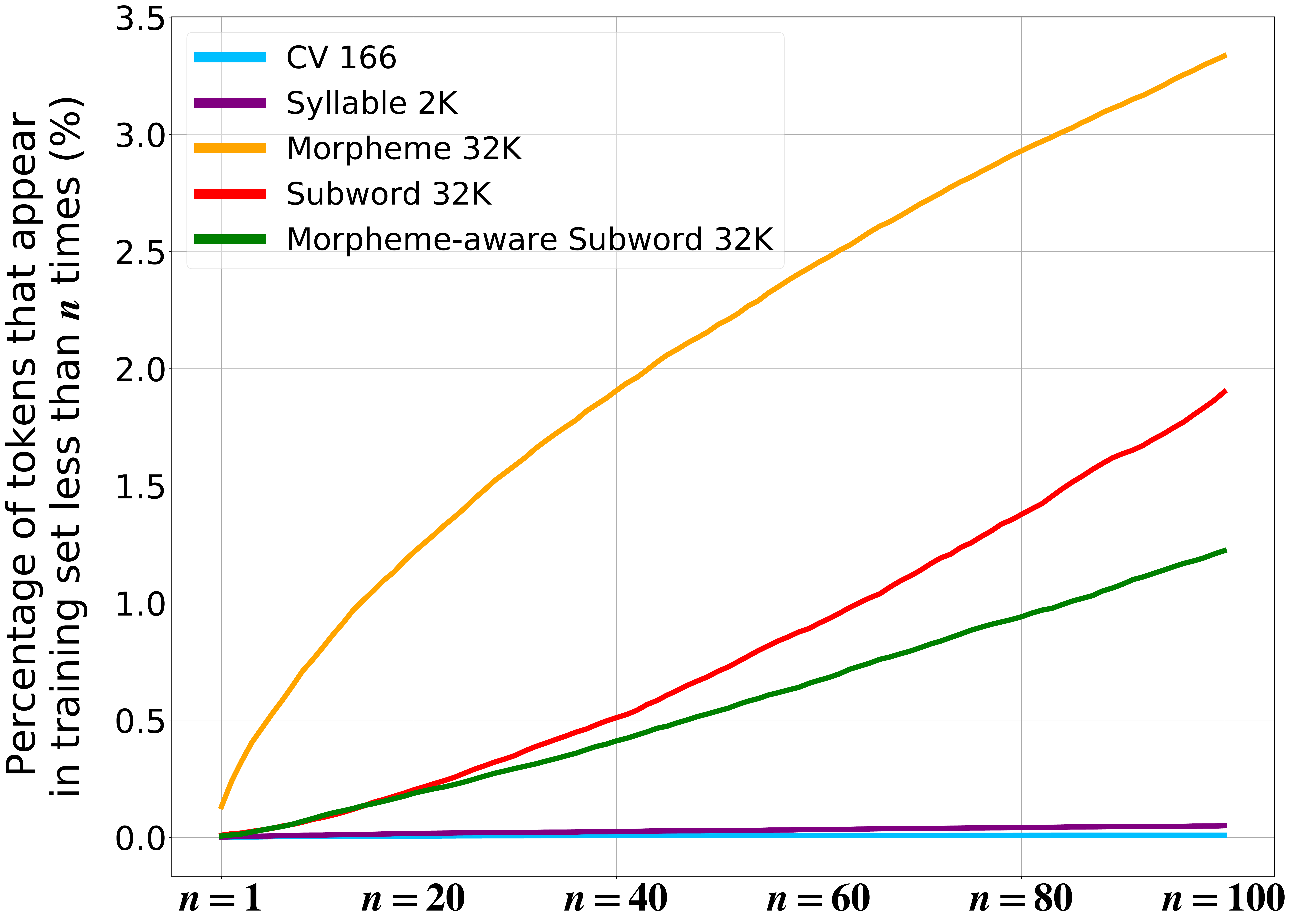}
    \caption{Percentage of under-trained tokens in various tokenization strategies}
    \label{fig:under_train}
\end{figure}

OOVs are the tokens that appear only in the test set. They are an extreme case of under-trained tokens---test set's tokens that appear in the training set for the limited number of times. Figure \ref{fig:under_train} shows how much under-trained tokens account for in each model, ranging from $n=1$ to $n=100$, where $n$ is the frequency of the under-trained tokens in the training set.
Clearly, the curve of the Morpheme 32K model is far above that of the others, indicating that it suffers from the problem of under-trained tokens the most.

\section{Conclusion}
\label{sec:conclusion}

We explored various Korean tokenization strategies on machine translation and five NLU tasks.
In machine translation Morpheme-aware Subword models with a vocabulary size worked best for both Korean to English and English to Korean settings.
By contrast, there was no single best tokenization strategy for the NLU tasks. 
Instead, Subword 64K models showed the best performance on KorQuAD, whereas Morpheme-aware Subword 64K models turned out to be optimal for the other KorNLI, KorSTS, NSMC, and PAWS-X tasks.

\section*{Acknowledgments}

We are grateful to the anonymous reviewers for their valuable comments. For pre-training models, we used Cloud TPUs provided by TensorFlow Research Cloud program.

\bibliography{aacl-ijcnlp2020}
\bibliographystyle{acl_natbib}

\end{document}